\documentclass[11pt,a4paper]{article}
\usepackage[hyperref]{emnlp2017}

\usepackage{newtxtext,newtxmath}
\usepackage{latexsym}

\usepackage[detect-all]{siunitx}
\usepackage{etoolbox}
\robustify\bfseries

\usepackage{graphicx}
\usepackage{subcaption}

\usepackage{array}
\newcolumntype{P}[1]{>{\centering\arraybackslash}p{#1}}
\usepackage{multirow}

\usepackage{pgfplots}
\pgfplotsset{compat=1.14}

\usepackage{microtype}

\usepackage{url}
\emnlpfinalcopy
\def\emnlppaperid{21}

\title{An Investigation into the Pedagogical Features of Documents}

\author{
  Emily Sheng{\rm ,}
  Prem Natarajan{\rm ,}
  Jonathan Gordon{\rm , and}
  Gully Burns \\
  USC Information Sciences Institute \\
  Marina del Rey, CA, USA \\
  \{ewsheng,\,pnataraj,\,jgordon,\,burns\}@isi.edu
}

\date{}

\begin{document}

% This disallows LaTeX from stretching the space between paragraphs. This results in pages where the columns don't align, i.e., they are not flush at the bottom.
%\setlength{\parskip}{0pt}

\maketitle

\begin{abstract}

Characterizing the content of a technical document in terms of its learning utility can be useful for applications related to education, such as generating reading lists from large collections of documents. We refer to this learning utility as the ``pedagogical value'' of the document to the learner. While pedagogical value is an important concept that has been studied extensively within the education domain, there has been little work exploring it from a computational, i.e., natural language processing (NLP), perspective. To allow a computational exploration of this concept, we introduce the notion of ``pedagogical roles'' of documents (e.g., \textit{Tutorial} and \textit{Survey}) as an intermediary component for the study of pedagogical value. Given the lack of available corpora for our exploration, we create the first annotated corpus of pedagogical roles and use it to test baseline techniques for automatic prediction of such roles.
% * <jgordon@isi.edu> 2017-07-13T20:49:39.043Z:
%
% > e.g., Tutorial and Survey
%
% My preference is to italicize these roles throughout the paper. Otherwise, capitalized, they feel a bit like Truth and Beauty.
%
% ^ <ewsheng@isi.edu> 2017-07-14T00:14:30.451Z.
% * <jgordon@isi.edu> 2017-07-13T20:46:39.469Z:
%
% > The main contributions of this paper are a novel annotated corpus in which documents are tagged with labels related to the learning utility of the document, and baseline results from document-classification-style experiments using the corpus.
%
% I suggest dropping the first sentence of the abstract as redundant. (I otherwise would have moved it to the end of the abstract, but I think the current last sentence says much the same thing.)
%
% ^ <jgordon@isi.edu> 2017-07-14T16:54:09.318Z.
\end{abstract}

\section{Introduction}
We define ``pedagogical value'' as the estimate of how useful a document is to an individual who seeks to learn about specific concepts described in the document. 
%Measuring learning utility implies there is a learner and something to be learned; in our work, pedagogical value must be defined between a learner and a document. This concept of ``personal'' and ``situational'' variables in the learning process is well established in the domain of education \cite{biggs1987student}. Personal variables include prior knowledge, abilities, and personality, and situational variables include subject matter and teaching method. % In broader terms, ``knowledge is better understood as emergent in activity and cannot be represented in a way that is independent of its context'' \cite{Recker2003}.
% * <jgordon@isi.edu> 2017-07-13T20:52:40.363Z:
%
% > Measuring learning utility implies there is a learner and something to be learned; in our work, pedagogical value must be defined between a learner and a document. This concept of ``personal'' and ``situational'' variables in the learning process is well-established in the domain of education \cite{biggs1987student}. Personal variables include prior knowledge, abilities, and personality, and situational variables include subject matter and teaching method.
% Is this worth including? It's very broad. You could merge with the next paragraph.
%
% ^ <ewsheng@isi.edu> 2017-07-14T06:12:06.379Z.
A computational task that operationalizes the concept of pedagogical value is generating an ordered reading list of documents that a learner can traverse in order to maximize understanding of a subject. When a professor manually constructs a reading list about a specific subject for a student, the professor incorporates substantial knowledge of the subject history and interdependencies with other related subjects. The student's background and the relative qualities of documents on similar subjects are also considered. Techniques for automatically generating reading lists should also consider the extent to which a learner will be able to learn from a particular document.

Previously, \citet{tang2009pedagogical} have studied the ``pedagogical value of papers'' in the context of paper recommendation. In their work, they define the multiple ``pedagogical values'' of a paper as the paper's overall ratings, popularity, degree of peer recommendation, learner gain in new knowledge, learner interest, and learner background knowledge. Other efforts on generating reading lists and document recommendation have focused on modeling concepts represented in documents \cite{jardine2014automatically}, modeling concept dependencies \cite{gordon16}, and user modeling \cite{Bollacker:1999:SAP:313238.313270}, but there appears to be very limited work on characterizing the learning utility between a learner and a document. The abstract nature of pedagogical value motivates us to identify explicit document features that are salient to pedagogical value. With graduate students as our target learners, we start with a simplified model of novice, intermediate, and advanced learners, and we focus on identifying pedagogical features of documents that could benefit different learners.
% * <jgordon@isi.edu> 2017-07-13T20:55:46.767Z:
%
% > documents that could benefit different learners.
%
% Ensure that the final paper doesn't have this line at the start of the next page by itself. I won't edit to fix this right now since any changes to the abstract or introduction could take care of this organically.
%
% ^ <ewsheng@isi.edu> 2017-07-14T00:44:28.554Z.

In our document annotation process, we collected annotations for the qualitative and largely objective judgments of categories that documents belong to: \textit{Tutorial}, \textit{Survey}, \textit{Software Manual}, \textit{Resource}, \textit{Reference Work}, \textit{Empirical Results}, and \textit{Other}. We identify the seven categories based on document objectives in presenting content, e.g., \textit{Tutorials} teach the reader step by step how to do something, \textit{Resource} papers point the reader to datasets and implementations. Motivated by the need to conceptually organize information to be pedagogically useful, we refer to documents with different objectives as fulfilling different ``pedagogical roles.'' In the rest of this paper, we will use the document category names to refer to the pedagogical roles.
% * <jgordon@isi.edu> 2017-07-13T21:02:36.960Z:
%
% > fulfilling
% I changed it from "being in", but "serving" would also sound good, I think.
%
% ^ <ewsheng@isi.edu> 2017-07-13T22:47:51.961Z.

Identifying important qualitative features of pedagogical value, such as the pedagogical role, gives a greater degree of interoperability and insight into how we can help students learn more effectively. Education research explains the distinction between declarative and functioning knowledge: the former is knowledge of content and the latter is knowledge of how to interpret and put the content to work \cite{biggs2011teaching}. To apply content, learners must first understand the content; this explains why a novice and an advanced learner trying to learn the same subject would seek out documents with different pedagogical roles. \textit{Tutorials}, \textit{Reference Works}, and \textit{Survey} papers are better introductions for a novice with no knowledge of a subject. In contrast, an expert would have enough background knowledge to dive right into advanced papers presenting state-of-the-art empirical results. Although pedagogical roles are not the same as pedagogical value, these pedagogical features offer some insight as a starting point for estimating learning utility. For our study, we collected annotations for over 1000 documents, which we document and make available for others to use.\footnote{\url{https://doi.org/10.6084/m9.figshare.5202424}}
% * <jgordon@isi.edu> 2017-07-13T21:04:29.717Z:
%
% > We will make the annotations available to the public.
%
% Change to Figshare URL / DOI
%
% ^ <ewsheng@isi.edu> 2017-07-14T05:36:43.336Z.

We also collected annotations for three ordinal-scale questions of document complexity and quality as an exercise to gauge the feasibility of the task despite its subjective nature. However, the resulting inter-annotator agreement results were too low to be meaningful. These results stress the importance of identifying more objective user and document features relevant to pedagogical value; in this initial work, we focus on document features.

Our contribution is twofold: We provide the first annotated corpus of pedagogical roles for the study of pedagogical value, and we present baseline classification results using state-of-the-art techniques for others to work with. Our goal is to establish a framework that can be extended to other domains, provide empirical results to validate our dataset and algorithms, and demonstrate the feasibility of the proposed role classification task. In the rest of this paper, we will describe our methods for collecting, evaluating, and automatically generating annotations in \autoref{sec:methods}, the results of our evaluations in \autoref{sec:results}, related work in \autoref{sec:related}, and concluding remarks in \autoref{sec:conclusion}.
% * <jgordon@isi.edu> 2017-07-13T21:06:47.454Z:
%
% > We provide the first annotated corpus of pedagogical roles for the study of pedagogical value through pedagogical features
%
% I know what you mean, but I think you need to find a way to say this without saying "pedagogical" three times in quick succession.
%
% ^ <ewsheng@isi.edu> 2017-07-13T22:51:06.746Z.
% * <jgordon@isi.edu> 2017-07-13T21:06:14.265Z:
%
% > classification results of state-of-the-art techniques
%
% Confusing phrasing.
%
% "classification results using state-of-the-art techniques"?
% "results of classification with state-of-the-art techniques"?
%
% ^ <ewsheng@isi.edu> 2017-07-13T22:52:56.582Z.

\section{Methods}
\label{sec:methods}

\subsection{Creating guidelines for annotation}
We performed a few rounds of annotation to develop a set of roles that would be adequate and insightful for an initial investigation. We identified the following pedagogical roles:

\begin{itemize}
\setlength\itemsep{0.2em}
\item
\textbf{\textit{Survey}:} Is this document a broad survey? A broad survey examines or compares across a broad concept.
\item
\textbf{\textit{Tutorial}:} Is this document a tutorial? \textit{Tutorials} describe a coherent process about how to use tools or understand a concept, and teach by example.
\item
\textit{\textbf{Resource}:} Does this document describe the authors' implementation of a system, corpus, or other resource that has been distributed (e.g., public data sets or tools that have been released under an open-source license or are commercially available)?
\item
\textbf{\textit{Reference Work}:} Is this document a collection of authoritative facts intended for others to refer to? Reports of novel, experimental results are not authoritative facts; the statement ``grass is green'' is. \textit{Reference Works} describe different subtopics within a concept.
\item
\textbf{\textit{Empirical Results}:} Does this document describe results of the authors' experiments?
\item
\textbf{\textit{Software Manual}:} Is this document a manual describing how to use different components of a software?
\item
\textbf{\textit{Other}:} Other role. This includes theoretical papers, papers that present a rebuttal for a claim, thought experiments, etc.
\end{itemize}

Additionally, we developed annotation guidelines instructing annotators to select all applicable pedagogical roles for each document. A document could present results of a novel method and also direct readers to an implementation of the method, thus making the paper both an \textit{Empirical Results} paper and a \textit{Resource} paper. Another document could simultaneously give a step-by-step tutorial about how to use a system, present specific commands on how to use components of the system, and provide a link to where readers can download the system, making the document a \textit{Tutorial}, \textit{Software Manual}, and \textit{Resource}. Although a document could validly belong to multiple pedagogical roles, we have carefully gone through several iterations of pedagogical roles to maximize the differences between roles. In other words, the distribution of the number of pedagogical roles per document is skewed such that most of the documents have one role. The \textit{Other} role is an alternative category for all other possible pedagogical role types; we do not focus on documents with this role in this work. We believe most of the \textit{Other} documents have high pedagogical value to a small group of experts and are beyond the scope of this initial investigation. In addition to these guidelines, we also provided a few examples of documents of each pedagogical role to annotators.

\subsection{Annotation}

The corpus of documents we annotated is drawn from a collection of pedagogically diverse documents related to natural language processing. The collection is based on the ACL Anthology, using the plain-text documents included in the ACL Anthology Network corpus \cite{Radev&al.09a}. The ACL Anthology primarily consists of expert-level empirical research papers, so the collection was expanded to include other document types, as described in \citet{GordonEtAl2017}. %For gathering additional documents, we identified a set of query terms by finding the Wikipedia page titles (e.g., ``Markov logic network'') that occur most often in the ACL Anthology and filtered out those that are not reasonable research terms. We then downloaded relevant articles from Wikipedia, book chapters from Elsevier ScienceDirect, and ``tutorial'' articles from searching the web. 
Although we generally targeted specific document sources for specific pedagogical roles, we still found a variety of pedagogical roles from each source, i.e., not all documents from Wikipedia are \textit{Reference Works}, and not all papers found while searching the web for ``tutorials'' are \textit{Tutorials}. For annotation, we tried to identify a balanced sample of documents with different roles in this corpus by using simple regular expression pattern matching in document titles and abstracts. For example, to roughly target \textit{Software Manuals}, we looked for documents with the phrase ``software manual,'' ``manual,'' or ``technical manual'' in the title or abstract.

To choose a reliable group of annotators, we internally annotated pedagogical roles for a set of documents and compared it with annotations done by a group of students pursuing master's degrees in computer science. We selected 11 students whose annotations had the highest correlation with our annotations. These annotators were instructed to read the abstract if there was one and to skim the rest of the document in enough detail such that they were able to annotate features for the document accurately and in a timely manner. We met regularly to discuss and come to a consensus on general document characteristics that were confusing to interpret.

We divided the documents for annotation into subsets of 100 to distribute among annotators so that each document was annotated by three annotators, and each subset was annotated by the same three annotators. We also manually filtered through and internally annotated 155 more supplementary documents to make up for a lack of documents that were annotated as \textit{Surveys}, \textit{Resources}, and \textit{Software Manuals}. This supplementary set consists of 76 documents from the expanded ACL corpus and 79 additional documents collected from searching the web for more \textit{Surveys}, \textit{Resources}, and \textit{Software Manuals}.\footnote{Supplementary annotations are included in our publicly available annotation dataset.}

\subsection{Automatic prediction of pedagogical roles}
We represent each document as a bag of sentence-embedding clusters. This technique embeds all sentences into vectors, clusters sentence vectors, and then represents documents as distributions over clusters. To evaluate the effectiveness of representing each document as a bag of sentence-embedding clusters and performing k-nearest neighbors classification, we also run two baseline techniques. One baseline technique is a multi-label centroid-based algorithm with sentence embeddings that is related to the single label centroid-based algorithm presented by \newcite{han2000centroid} and the na\"\i ve \newcite{rocchio1971relevance} classification algorithm, a popular method for text classification \cite{rogati2002high}. The other baseline technique is a random forest classification of TF--IDF scores, which allows us to evaluate if sentence embeddings are more useful than word frequencies for this task. 

We use sentence embeddings because specific sentences in documents are key indicators of the pedagogical roles of the document. As an explicit example, one might find the following in a \textit{Survey} paper: ``This paper presents a survey of the field of machine translation\ldots'' A more implicit example might be a \textit{Resource} paper that mentions that one can find the corpus created by the authors at a specific link. We want to give much weight to the sentences that are the best indicators of the pedagogical roles of the document and leverage this information to automatically predict the pedagogical roles of documents. Skip-thought vectors\footnote{\url{https://github.com/ryankiros/skip-thoughts}} are able to effectively capture the semantics and syntax of sentences in several different tasks \cite{kiros2015skip}. To generate sentence embeddings needed for the centroid-based algorithm and the bag of sentence embedding clusters, we apply skip-thought vectors to embed each sentence from our annotated documents into a 4800-dimensional vector. We use the pre-trained skip-thought vector model to create sentence embeddings for each sentence.\footnote{Model parameter details in Supplemental Material \ref{sec:skip-thought-parameters}.}

In our techniques, we do not pre-select sentences to include as features for classifying a document. We also do not treat sentences differently given their location in different sections of a document, e.g., introduction versus conclusion. Our corpus is composed of research papers, book chapters, Wikipedia articles, and web documents, so there is not a standard format that all documents follow. Our goal is to discover different types of sentences that could support our defined set of pedagogical roles as well as point to the existence of other roles.

% \paragraph{Simple baseline classifier (SIM):} For each pedagogical role, we manually create a list of phrases (Table \ref{table:simple}) that would indicate whether a document belongs to the pedagogical role. We do not think there are any particular phrases that are commonly used to introduce Reference works, so we instead classify all documents from Wikipedia and Elsevier's ScienceDirect as Reference works.

% \begin{table}[th]
% \resizebox{!}{0.12\textheight}{
% \centering
% \begin{tabular}{|p{2.2cm}|p{4.7cm}|}
% \hline
% Pedagogical role & Keyphrases/Criteria \\
% \hline
% Survey & ``survey''\\
% \hline
% Tutorial & ``tutorial''  \\
% \hline
% Resource & ``publicly available'' \\
% \hline
% Ref. work & whether the document source is Wikipedia or Elsevier ScienceDirect \\
% \hline
% Emp. results & ``empirical'', ``results'', ``score'' \\
% \hline
% Sof. manual & ``software manual'', ``manual'', ``technical manual'' \\
% \hline
% Other & documents that don't belong to any other pedagogical role \\
% \hline
% \end{tabular}
% }
% \caption{Keyphrases or other criteria used to classify different pedagogical roles}
% \label{table:simple}
% \end{table}

% To classify each document, we look for each keyphrase for each pedagogical role in the first five sentences of the document\footnote{except Reference works and Other papers}, where titles are counted as the first sentence. For each pedagogical role, if any keyphrase of the role is present in the first five sentences, the document is classified as the pedagogical role.

\paragraph{Random Forest baseline classifier (RF):} TF--IDF scores of words in our annotated documents are used as features for a random forest classifier. To calculate the TF--IDF scores, we included words that were in at least 10\% and at most 90\% of the documents. We used five-fold cross-validation to evaluate the results.\footnote{Model parameter details in Supplemental Material \ref{sec:random-forest-parameters}.}

\paragraph{Multi-label centroid-based algorithm with sentence embeddings (CEN):} Each pedagogical role is represented by an average centroid vector, which is calculated by adding all sentence vectors in every document that belongs to the role, and then dividing the sum by the total number of sentence vectors added. When classifying a new document, we assign each sentence vector in the new document to a role label based on the nearest average vector. The role labels that are predicted for more than a third of the document's sentences are then predicted to be the document's role(s). Although this baseline method limits each document to two or fewer role predictions, it works as a rough baseline. 99.1\% of the annotated documents have one or two pedagogical roles, and we assume our sample of annotated documents is representative of a larger collection of documents.

\begin{figure*}[!ht]
 \includegraphics[width=\linewidth]{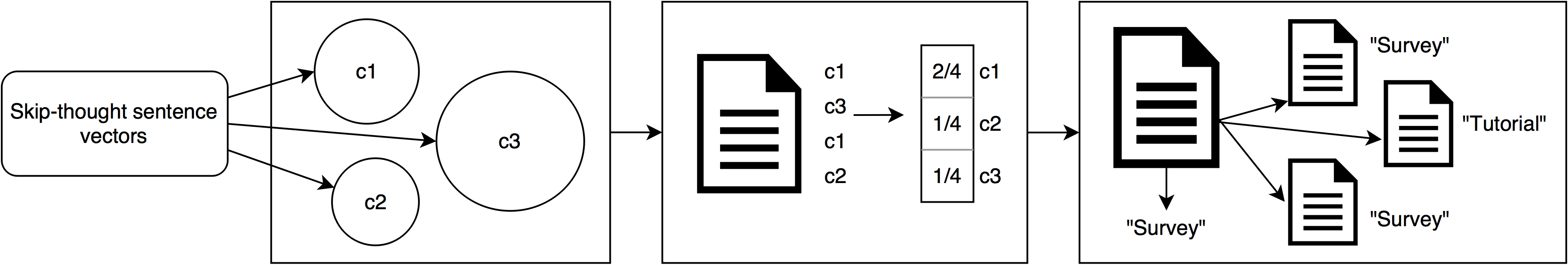}

  \begin{minipage}[t]{0.15\linewidth}
    \centering
    \subcaption{}\label{sentencevectors}
  \end{minipage}
  \begin{minipage}[t]{.25\linewidth}
    \centering
    \subcaption{}\label{sentenceclustering}
  \end{minipage}
  \begin{minipage}[t]{.27\linewidth}
    \centering
    \subcaption{}\label{clusterdistribution}
  \end{minipage}
  \begin{minipage}[t]{.3\linewidth}
    \centering
    \subcaption{}\label{knn}
  \end{minipage}

  \caption{An end-to-end overview of the BoSEC+KNN technique. In (a), we generate skip-thought sentence vectors for every sentence in all documents. We partition all sentence vectors into clusters in (b). In (c), we represent each document as a distribution over the clusters formed in (b). (d) shows the KNN pedagogical role classification of documents based on the majority votes of annotated documents.}
  \label{fig:docclustering}
\end{figure*}

\paragraph{Bag of Sentence Embedding Clusters (BoSEC):} Starting with the hypothesis that semantic and syntactic features of sentences are useful indicators of pedagogical roles, we employ $k$-means clustering\footnote{\url{http://scikit-learn.org}, model parameter details in Supplemental Material \ref{k-means-parameters}.} over sentence vectors to generate a representation basis (of $N$ clusters) for computing a single $N \times 1$ feature vector per document. Each entry in the feature vector is the relative frequency of the specific sentence vector cluster being observed in the document.

\paragraph{K-Nearest Neighbors with Bag of Sentence Embedding Clusters (KNN+BoSEC):} We use k-nearest neighbors classification to search for documents which exhibit the most similar distributions of clusters and predict the pedagogical roles of documents. To predict the roles of document $A$, we look for the three nearest documents in the $N$-dimensional vector space as calculated by the Manhattan distance metric. The majority roles of the three nearest documents are then predicted to be the roles of document $A$. The details of KNN+BoSEC are shown in \autoref{fig:docclustering}.

\paragraph{KNN+BoSEC with custom sentence encoder (KNN+BoSEC+):} The content and style of writing in the scientific papers in our corpus differs from that of books used to train the pre-trained skip-thoughts vector model. We also run experiments using the KNN+BoSEC technique with a custom sentence embedding model trained on our entire collection of (annotated and unannotated) NLP documents. The custom sentence embedding model is trained using the default parameters described in the skip-thoughts training code.\footnote{\url{https://github.com/ryankiros/skip-thoughts}; model parameter details in Supplemental Material \ref{custom-skip-thought-parameters}.}

\section{Results}
\label{sec:results}

\subsection{Annotation agreement evaluation}
The kappa value, which measures the likelihood of annotator agreement occurring above chance, is 0.68 for the pedagogical role annotations. This kappa value was calculated as an average over the kappa values for each subset of 100 documents. Given the difficulty of annotating pedagogical roles, which was confirmed by annotators, we believe a kappa of 0.68 indicates substantial agreement between annotators \cite{landis1977measurement}.

\autoref{table:interannotatorconfusion} shows the details of inter-annotator agreement for annotated pedagogical roles from documents with only one majority role. The rows are the majority roles, which we take to be the ground truth pedagogical roles of documents. The columns show the third annotator's annotations; if the third annotation matches the majority, then the particular annotation falls on the diagonal of \autoref{table:interannotatorconfusion}. Although there are 1264 majority pedagogical role annotations, we calculated the confusion matrix for 1206 roles from documents with only one majority role each, for ease of interpretation. From the 1206 pedagogical roles, there are 1245 role pairs between the majority role and the third annotator's annotated role(s).

\begin{table}[th]
\centering
\footnotesize
\def\arraystretch{1.4}
\sisetup{table-format=2}
\begin{tabular}{@{}l|SSSS[table-format=3]S[table-format=3]SS[table-format=3]|S[table-format=4]@{}}
\hline
&
{\rotatebox[origin=l]{90}{\textit{Survey}}} &
{\rotatebox[origin=l]{90}{\textit{Tutorial}}} &
{\rotatebox[origin=l]{90}{\textit{Resource}}} &
{\rotatebox[origin=l]{90}{\textit{Reference Work}}} &
{\rotatebox[origin=l]{90}{\textit{Empirical Results} }} &
{\rotatebox[origin=l]{90}{\textit{Software Manual}}} &
{\rotatebox[origin=l]{90}{\textit{Other}}} &
{\rotatebox[origin=l]{90}{Total}} \\
\hline
\textit{Sur.} & \bfseries 10 & 1 & 0 & 7 & 4 & 0 & 5 & 27 \\
\textit{Tut.} & 2 & \bfseries 44 & 6 & 22 & 6 & 4 & 14 & 98  \\
\textit{Res.} & 0 & 0 & \bfseries 5 & 1 & 1 & 3 & 5 & 15 \\
\textit{Ref.} & 36 & 20 & 3 & \bfseries 151 & 4 & 1 & 28 & 243 \\
\textit{Emp.} & 13 & 8 & 8 & 15 & \bfseries 526 & 3 & 56 & 629 \\
\textit{Sof.} & 0 & 1 & 0 & 0 & 2 & \bfseries 1 & 2 & 6 \\
\textit{Other} & 12 & 24 & 6 & 47 & 29 & 2 & \bfseries 107 & 227 \\
\hline
Total & 73 & 98 & 28 & 243 & 572 & 14 & 217 & \bfseries 1245 \\
\hline
\end{tabular}
\caption{Confusion matrix for annotated pedagogical roles from documents with only one majority role. Rows are the majority roles (chosen by two or three annotators) that we treat as ground truth. Columns are the third annotator's corresponding annotations.}
\label{table:interannotatorconfusion}
\end{table}

From \autoref{table:interannotatorconfusion}, we can see that \textit{Survey} documents are sometimes confused with \textit{Reference Works}, \textit{Resource} papers are sometimes confused with \textit{Other} documents, and \textit{Software Manuals} are rare. We also see that \textit{Other} documents have relatively higher rates of misclassification. These results are consistent with feedback from annotators. The reason why \textit{Survey} documents are sometimes mistaken for \textit{Reference Works} is because both examine a broad number of subjects in a domain; the distinction we make in our annotation guidelines is that \textit{Reference Works} are a collection of established authoritative facts such as those one might find in an encyclopedia, whereas \textit{Surveys} focus on the discoveries of other publications. When looking for \textit{Resource} papers, annotators rely on looking for few indicator sentences that may be missed with a more superficial skim of the document. Also, the \textit{Other} documents belong to a range of additional pedagogical roles, though we do not make finer distinctions here.

For each annotated document, we kept the pedagogical roles that had majority annotation agreement across the three annotators who annotated the document. If a document had no majority labels, the document was filtered out of the annotated document set. This filtered document set of 1235 documents with 1264 annotated pedagogical roles is the one we use along with a supplementary set for all pedagogical role prediction techniques.

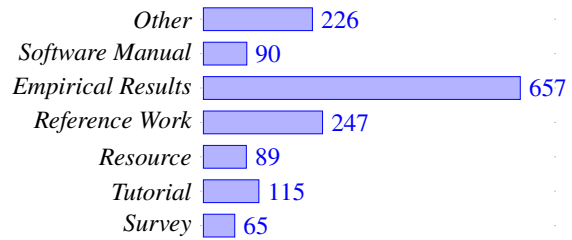
\begin{figure}[ht]
\resizebox{0.48\textwidth}{!}{
    \begin{tikzpicture}
      \begin{axis}[
        xbar,
        font = \itshape,
        y axis line style = { opacity = 0 },
        axis x line       = none,
        width 			  = 7cm,
        tickwidth         = 0pt,
        ytick 			  = data,
        enlarge y limits  = 0.2,
        enlarge x limits  = 0.12,
        symbolic y coords = {Survey, Tutorial, Resource, Reference Work, Empirical Results, Software Manual, Other},
        nodes near coords
      ]
      \addplot coordinates { (226,Other) (90,Software Manual)
      (657,Empirical Results) (247,Reference Work)
      (89,Resource) (115,Tutorial) (65,Survey) };
      \end{axis}
    \end{tikzpicture}
    }
    \caption{Distribution of all pedagogical role annotations in the full annotated corpus used for training classifiers}
  	\label{fig:corpusstats}
\end{figure}

We noticed a lack of \textit{Surveys}, \textit{Resources}, and \textit{Software Manuals}, so we internally annotated another supplementary set of 155 documents consisting mostly of documents of the underrepresented roles. The full annotated corpus we use for classification has the distribution of roles shown in \autoref{fig:corpusstats}; this full corpus includes the filtered set of 1235 documents annotated by three annotators each and 155 internally annotated documents, for a total of 1489 pedagogical role annotations over 1390 documents. Given the corpora we selected our set of documents to annotate from, it is not surprising that most of the documents are \textit{Empirical Results}, \textit{Reference Works}, \textit{Tutorials}, or \textit{Other}. 94\% of the annotated documents have just one pedagogical role, and 99.1\% have one or two pedagogical roles.\footnote{See \autoref{fig:numannotations} in Supplemental Material for more details.} The top three most common combinations of roles for a document are \textit{Resource} and \textit{Empirical Results}; \textit{Resource} and \textit{Software Manual}; \textit{Tutorial}, \textit{Resource}, and \textit{Software Manual}.\footnote{See \autoref{fig:multipleroles} in Supplemental Material for more details.} Many documents with multiple pedagogical roles are \textit{Resource} documents because the authors make their work publicly available.

\subsection{Pedagogical role classification evaluation}
\begin{table*}[th]
\sisetup{table-format=1.2}
\def\arraystretch{1.2}
\setlength{\tabcolsep}{0.45em}
\centering
\begin{tabular}{@{}P{2cm} | S S S S | S S S S | S S S S | S[table-format=3.1]}
\hline
\multirow{2}{*}{Ped.\ role} &
\multicolumn{4}{c|}{Precision} &
\multicolumn{4}{c|}{Recall} &
\multicolumn{4}{c|}{$F_1$} &
Support \\ [0.5ex]
&
{\rotatebox[origin=l]{90}{RF}} &
{\rotatebox[origin=l]{90}{CEN}} &
{\rotatebox[origin=l]{90}{KNN+BoSEC}} &
{\rotatebox[origin=l]{90}{KNN+BoSEC+}} &
{\rotatebox[origin=l]{90}{RF}} &
{\rotatebox[origin=l]{90}{CEN}} &
{\rotatebox[origin=l]{90}{KNN+BoSEC}} &
{\rotatebox[origin=l]{90}{KNN+BoSEC+}} &
{\rotatebox[origin=l]{90}{RF}} &
{\rotatebox[origin=l]{90}{CEN}} &
{\rotatebox[origin=l]{90}{KNN+BoSEC}} &
{\rotatebox[origin=l]{90}{KNN+BoSEC+}} &
{\rotatebox[origin=l]{90}{All}} \\
\hline
\textit{Survey} &
0 & 0.02 & 0.23 & \textbf{0.31} &
0 & \textbf{0.21} & 0.20 & 0.18 &
0 & 0.03 & 0.21 & \textbf{0.23} & 13 \\
\textit{Tutorial }&
0.50 & 0.10 & 0.64 & \textbf{0.66} &
0.05 & 0.21 & \textbf{0.55} & 0.52 &
0.08 & 0.11 & 0.57 & \textbf{0.58} & 23 \\
\textit{Resource} &
0.20 & 0 & \textbf{0.70} & 0.53 &
0.01 & 0 & 0.19 & \textbf{0.24} &
0.03 & 0 & 0.29 & \textbf{0.32} & 17.8 \\
\textit{Ref.\ Work} &
0.77 & 0.07 & 0.71 & \textbf{0.78} &
0.33 & 0.32 & 0.70 & \textbf{0.71} &
0.46 & 0.11 & 0.70 & \textbf{0.74} & 49.4 \\
\textit{Emp.\ Res.} &
\textbf{0.86} & 0 & 0.83 & 0.85 &
0.77 & 0 & 0.86 & \textbf{0.89} &
0.81 & 0 & 0.85 & \textbf{0.87} & 131.4 \\
\textit{Sof.\ Man.} &
\textbf{0.98} & 0.05 & 0.93 & 0.95 &
0.34 & 0.16 & 0.72 & \textbf{0.86} &
0.49 & 0.07 & 0.81 & \textbf{0.90} & 18 \\
\textit{Other} &
0.63 & 0.06 & 0.57 & \textbf{0.65} &
0.10 & 0.40 & 0.27 & \textbf{0.48} &
 0.17 & 0.10 & 0.36 & \textbf{0.55} & 45.2 \\
\hline
avg / total &
0.71 & 0.03 & 0.73 & \textbf{0.76} &
0.44 & 0.15 & 0.64 & \textbf{0.70} &
0.50 & 0.05 & 0.66 & \textbf{0.72} & 297.8 \\
\hline
\end{tabular}
\caption{Precision, recall, and $F_1$ scores by pedagogical roles for all methods. Support is the actual number of documents with each role. avg / total computes weighted averages of scores across all roles. All values are averaged over a five-fold cross validation.}
\label{table:rolereport}
\end{table*}

In \autoref{table:rolereport}, we see that for both random forest classification of TF--IDF scores (RF) and sentence embedding methods (CEN, KNN+BoSEC, KNN+BoSEC+), the more samples there are for a pedagogical role, the higher the scores are for the role. The scores for \textit{Other} documents are an anticipated exception to the trend, because we do not make more fine-grained distinctions between other pedagogical roles in this work. \textit{Software Manuals} are also an exception to this trend, as their scores are relatively high for the number of samples; this is because \textit{Software Manuals} are typically written in a very distinct style. CEN generally performs poorly across roles, doing worse than the baseline random forest classification with TF--IDF. This suggests that word frequency is more informative about the pedagogical roles of a document than a single representative vector per role.

With the exception of \textit{Software Manuals}, RF is able to predict roles with more samples (\textit{Reference Work}, \textit{Empirical Results}, \textit{Other}) with higher precision compared to roles with fewer samples (\textit{Survey}, \textit{Tutorial}, \textit{Resource}). KNN+BoSEC and KNN+BoSEC+ have comparable precision for roles with more samples, but have significantly higher precision for roles with fewer samples. Compared to RF, KNN+BoSEC and KNN+BoSEC+ also have higher recall across all roles. KNN+BoSEC+ has the highest $F_1$ scores for all pedagogical roles. We attribute the fact that KNN+BoSEC+ is generally able to do better than KNN+BoSEC to using a custom sentence encoder trained on scientific documents.

% Paragraph below added to address feedback of comparing sentence-based methods with something simpler and keyword-based
Given that we use keyphrases to find documents that likely belong to specific pedagogical roles, we also want to see if we could achieve performance similar to that of our sentence embedding-based methods by simply classifying documents based on keyphrases. We manually curate a list of keyphrases for two pedagogical roles: ``software manual,'' ``manual,'' and ``technical manual'' for \textit{Software Manuals}, and ``tutorial'' for \textit{Tutorials}. We then classify a document as a certain role if any of the role's keyphrases are present in the first five sentences of the document, where the title counts as the first sentence. Classifying \textit{Software Manuals} with this method has a precision of 0.15, a recall of 0.09, and an $F_1$ score of 0.11. KNN+BoSEC+ dramatically outperforms this method with the specified keyphrases for \textit{Software manuals}. Classifying \textit{Tutorials} with this method has a precision of 0.60, a recall of 0.50, and an $F_1$ score of 0.55. While the keyphrase classification results for \textit{Tutorials} are closer to the corresponding KNN+BoSEC+ results, we think that the KNN+BoSEC+ results would also improve if it had access to the list of keyphrases as features, though we leave that for future experimentation. These initial keyphrase classification experiments suggest that sentence-embedding-based methods are generally more effective and robust than handcrafting keyphrases for each pedagogical role.

% KNN+BoSEC+ outperforms the SIM baseline on all pedagogical roles except Surveys and Reference works. The SIM baseline classifies all documents from Wikipedia and Elsevier ScienceDirect as Reference works, so although this baseline more accurately classifies documents as Reference works, this method is not generalizable for documents from unseen sources. The SIM results show that authors generally use the term ``survey'' in their survey papers, and that sentence embedding-based methods are still generally more effective at this classification task.

\begin{table}[th]
\hspace{-1em}
\resizebox{!}{0.13\textheight}{
\footnotesize
\sisetup{table-format=1.1}
\setlength{\tabcolsep}{0.4em}
\def\arraystretch{1.4}
\begin{tabular}{@{}l | S S[table-format=2.1] S S[table-format=2.1] S[table-format=3.1] S[table-format=2.1] S[table-format=2.1] S[table-format=2.1] | S[table-format=3.1]@{}}
\hline
&
{\rotatebox[origin=l]{90}{\textit{Survey} }} &
{\rotatebox[origin=l]{90}{\textit{Tutorial} }} &
{\rotatebox[origin=l]{90}{\textit{Resource} }} &
{\rotatebox[origin=l]{90}{\textit{Reference Work} }} &
{\rotatebox[origin=l]{90}{\textit{Empirical Results} }} &
{\rotatebox[origin=l]{90}{\textit{Software Manual} }} &
{\rotatebox[origin=l]{90}{\textit{Other} }} &
{\rotatebox[origin=l]{90}{No prediction }} &
{\rotatebox[origin=l]{90}{Total }} \\
\hline
\textit{Sur.} & \bfseries 2 & 0.2 & 0 & 0.8 & 4.2 & 0 & 1.2 & 1.6 & 10 \\
\textit{Tut.} & 0.2 & \bfseries 9.4 & 0 & 2.6 & 0.8 & 0.8 & 2 & 2.2 & 18 \\
\textit{Res.} & 0.2 & 0 & \bfseries 1.2 & 0 & 3 & 0 & 0.6 & 1.8 & 6.8 \\
\textit{Ref.} & 1.2 & 1.6 & 0.2 & \bfseries 34 & 3.6 & 0.2 & 3.2 & 4.2 & 48.2 \\
\textit{Emp.} & 0.8 & 1.4 & 1.8 & 3 & \bfseries 109.2 & 0 & 4 & 4.2 & 124.4 \\
\textit{Sof.} & 0 & 1.6 & 0.6 & 0.2 & 0 & \bfseries 9.8 & 0 & 0.4 & 12.6 \\
\textit{Oth.} & 3.4 & 0.6 & 0.6 & 3.2 & 8 & 0 & \bfseries 21.8 & 7.6 & 45.2 \\
\hline
Tot. & 7.8 & 14.8 & 4.4 & 43.8 & 128.8 & 10.8 & 32.8 & 22 & 265.2 \\
\hline
\end{tabular}
}
\caption{Ground truth pedagogical roles (rows) versus predicted roles (columns) using KNN+BoSEC+. We calculate the confusion matrix for documents with only one ground truth role. All values are averaged over a five-fold cross validation.}
\label{table:roleconfusionmatrix}
\end{table}

The confusion matrix in \autoref{table:roleconfusionmatrix} allows us to make judgments about documents of different pedagogical roles, as predicted by KNN+BoSEC+. The rows are the ground truth roles, and the columns are the predicted roles. We can see that \textit{Surveys}, \textit{Resources}, and \textit{Other} documents are often mistaken to be documents with \textit{Empirical Results}. Additionally, there are relatively more instances of \textit{Surveys}, \textit{Resources}, and \textit{Other} documents where the classifier is unable to make a prediction. Overall, these results suggest that the misclassifications are an effect of an unbalanced dataset with many more samples of \textit{Empirical Results}, rather than an inherent lack of distinctness between documents of different roles.

Through a qualitative analysis of sentences from the clusters most frequently associated with each pedagogical role, we observe that example sentences from different roles align with our intuitions of what exemplary sentences from different roles should be. The \textit{Survey} sentences describe progress in different areas of research; the \textit{Tutorial} sentences explain details of specific concepts and methods; the \textit{Software Manual} sentences give information about how to use a tool.\footnote{For more details, see \autoref{table:roleexamples} in Supplemental Material.} Sentences from the most frequent clusters of a role do not explicitly mention the roles of the paper, e.g., ``This paper presents a tutorial\ldots'' This phenomenon makes sense for two reasons. One reason is that the majority of documents do not explicitly say what kind of document they are. The second reason is that even when documents do explicitly state their role, the actual content of the document may disagree with the declared role. For example, some papers are written to accompany tutorials presented at workshops. The papers will explicitly declare themselves to be tutorials, but the paper will only include an abstract and not the tutorial itself. Following our annotation guidelines, we do not label these documents as \textit{Tutorials}. This implicit characterization of a document's pedagogical roles through sentences means that a method that merely searches for explicit mentions of keywords or declaration of the document's roles would not be an effective approach to this problem. Thus, these example sentences qualitatively validate our embedding and clustering approach to pedagogical role classification.

\section{Related Work}
\label{sec:related}

% Research at the intersection of natural language processing (NLP) and education generally use natural language processing techniques to do one of the following: teach language, use language to teach, and organize and facilitate the retrieval of content knowledge (Litman, 2016). Areas of research that use NLP techniques to teach language include grammar correction and essay scoring (Lee and Seneff, 2006; Shermis and Burstein, 2013). An example of a research area that uses languages to teach is intelligent tutoring systems in specialized domains (Murray, 1999). The third category of research at the junction of NLP and education includes recommendation systems for learning materials and the generation of reading lists (Tang and McCalla, 2004; Gori and Pucci, 2006; Santos and Boticario, 2010).

To the best of our knowledge, there is not much prior work that is directly related to investigating relevant pedagogical features of documents through pedagogical roles. There are some document recommendation systems that try to find documents that are both conceptually relevant to a user's query and pertinent to the user's interest, level of background knowledge, etc. For example, Semantic Scholar\footnote{\url{https://www.semanticscholar.org}} allows users to filter an automatically generated reading list by ``overviews,'' which are analogous to our definition of Surveys. PageRank accounts for popularity when identifying documents of interest \cite{page1999pagerank}. \citet{tang2004pedagogically} consider the user's background knowledge, interest towards specific topics, and motivation when making recommendations. \citet{gori2006research} present a research paper recommender system based on the random walk algorithm and a small set of papers that users mark as relevant. \citet{santos2010modeling} emphasize that recommendation systems in the e-learning domain should be ``guided by educational objectives'' and define a semantic model for recommendation objects.

Previous efforts at investigating the value of documents include evaluating the reading difficulty of documents, citation graphs, and surveys, though none really address the problem of estimating the pedagogical value of a document to a learner while focusing on the interpretability of the results. The interpretability of results is especially important in education because educators need to be able to provide clear feedback to students. In automatic essay scoring, researchers look at features such as word count, semantic and syntactic coherence, sentence length, vocabulary complexity, and the use of certain phrases that facilitate the flow of ideas, e.g., ``first of all'' \cite{burstein2004automated,shermis2013handbook}. These features are a starting point to estimate the value of a document, but to estimate pedagogical value, we must consider if and how these features would affect different learners. Other directions of research use the influence of a paper within a citation graph as a proxy for the value of the paper, following the reasoning that good quality papers would be more important ``nodes'' in a citation graph \cite{ekstrand2010automatically}; however, documents that are important ``nodes'' in the graph do not necessarily have high pedagogical value for all learners. \citet{tang2009pedagogical} present surveys to students as an annotation method to estimate the value of the paper to the learner. %``degree of difficulty to understand [the paper's] degree of job-relatedness with the user, its interestingness, its degree of usefulness, its ability to expand the user's knowledge (value-added), and its overall rating.''
They annotate individual features of job-relatedness, interestingness, usefulness, etc., using ordinal-scale values, and study the partial correlations between features to analyze the composition of features that contribute to the pedagogical value of a document. Our approach is different in that (a) we develop an intermediate representation of pedagogical value that can be largely objectively annotated, (b) we evaluate correlation between annotators and not between features, and (c) we additionally present baseline results of pedagogical role prediction.

The classification task described in this work is also related to text classification, a task with a long history in NLP. \citet{sebastiani2002machine} presents a detailed survey of tasks and techniques used in text classification up until the early 2000s. \citet{joachims1998text} presents experimental results that justify the use of Support Vector Machines (SVMs) for text classification. \citet{soucy2001simple} use TF--IDF scores and a KNN model to perform different text categorization tasks.

\section{Conclusion}
\label{sec:conclusion}

In this paper, we have described (a) our creation of the first annotated corpus of pedagogical roles for the study of pedagogical value and (b) our use of sentence embeddings and clustering techniques to develop a baseline for pedagogical role classification. The inter-annotator agreement for the annotation of pedagogical roles is substantial and thus a good basis to develop pedagogical role classification techniques and intuitions about pedagogical value upon. Analyses of our bag of sentence-embedding clusters technique support our intuition that certain sentences in a document are strong indicators of the pedagogical roles of the document. The next steps are to expand the set of roles as needed and apply our techniques to other domains in order to work towards a general approach to estimating pedagogical value. We believe it is important to make our corpus and annotations public, as feedback from other researchers will help improve the quality and scope of our corpus as we expand it. 

\section*{Acknowledgments}
The authors thank Yigal Arens, Aram Galstyan, and Linhong Zhu for their valuable feedback on this work.

This research is based upon work supported in part by the Office of
the Director of National Intelligence (ODNI), Intelligence Advanced
Research Projects Activity (IARPA), via Air Force Research Laboratory
(AFRL). The views and conclusions contained herein are those of the
authors and should not be interpreted as necessarily representing the
official policies or endorsements, either expressed or implied, of
ODNI, IARPA, AFRL, or the U.S. Government. The U.S. Government is
authorized to reproduce and distribute reprints for Governmental
purposes notwithstanding any copyright annotation thereon.

\bibliography{biblio}
\bibliographystyle{emnlp2017}

\appendix

\section{Supplemental Material}
\label{sec:supplemental}

\subsection{Skip-thought vector parameters}
\label{sec:skip-thought-parameters}

Each sentence vector has 4800 dimensions, with the first 2400 dimensions as the uni-skip model, and the latter 2400 dimensions as the bi-skip model. The model has the following parameters: recurrent matrices initialized with orthogonal initialization, non-recurrent matrices initialized from a uniform distribution in [$-0.1$, $0.1$], mini-batches of size 128, gradients clipped when the norm of the parameter vector is greater than 10, and the Adam algorithm for optimization.

\subsection{Random forest classification parameters}
\label{sec:random-forest-parameters}

For the random forest classifier, we used the Gini impurity function to estimate the quality of splits. When looking for the best split, the classifier considers the square root of the total number of features. The maximum depth of the tree is 75, and the classifier splits on a minimum of 5 samples at the internal nodes. We use 10 trees and a minimum of 1 sample at each leaf node.

\subsection{Mini-batch K-means parameters}
\label{k-means-parameters}

In this clustering technique, random subsets of the feature vectors are used in each iteration. We train the model with 300 clusters, early stopping if there is no improvement in the last 50 mini batches, a mini batch size of 4800, and the fraction of the maximum number of counts for a cluster center to be reassigned is 0.0001. We had experimented with different cluster sizes, and found 300 clusters to be the right size to maintain coherency within and distinction across clusters.

\subsection{Custom skip-thought vector model parameters}
\label{custom-skip-thought-parameters}

Specifically, the RNN word embeddings have 620 dimensions, and we use a uni-skip model with a hidden state size of 2400. Both the encoder and the decoder are GRUs. The size of the decoder vocabulary is 20000, and the maximum length of a sentence is 30 words; additional words in sentences are ignored. Our custom model is trained for 5 epochs, has a gradient clipping value of 5, has a batch size of 64, and uses the Adam optimization algorithm.

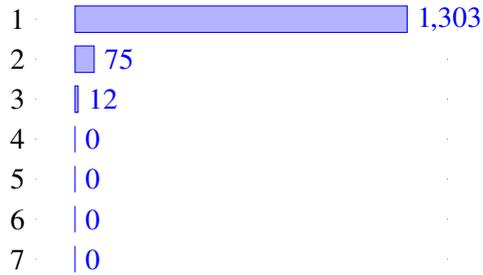
\begin{figure*}[ht]
	\centering
    \begin{tikzpicture}
      \begin{axis}[
        xbar,
        y axis line style = { opacity = 0 },
        axis x line       = none,
        tickwidth         = 0pt,
        width 			  = 7cm,
        ytick 			  = data,
        enlarge y limits  = 0.2,
        enlarge x limits  = 0.12,
        symbolic y coords = {7, 6, 5, 4, 3, 2, 1},
        nodes near coords
      ]
      \addplot coordinates { (1303,1) (75,2)
      (12,3) (0,4)
      (0,5) (0,6) (0,7) };
      \end{axis}
    \end{tikzpicture}
    \caption{Distribution of number of pedagogical roles per document in full annotated corpus}
  	\label{fig:numannotations}
\end{figure*}

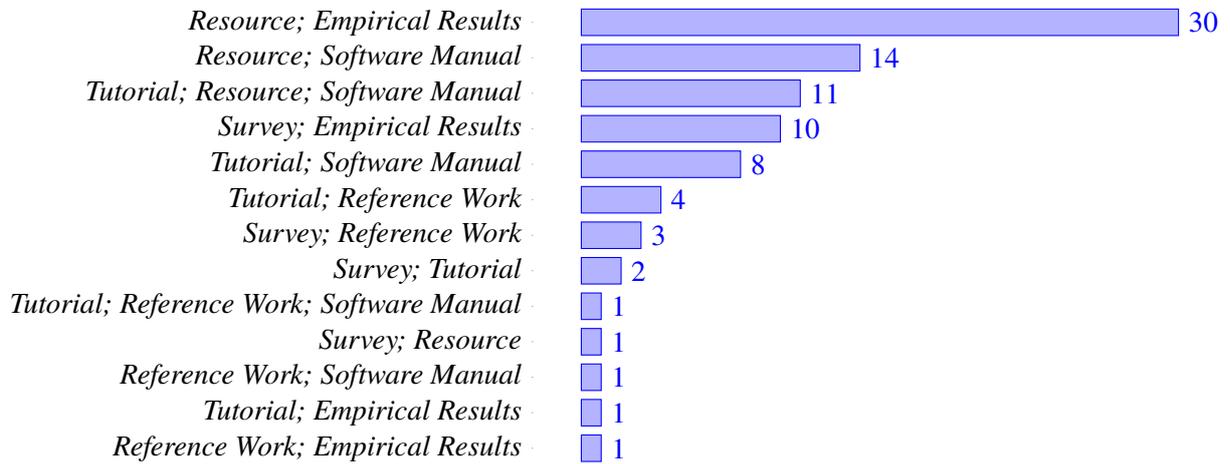
\begin{figure*}[ht]
			\begin{tikzpicture}
              \begin{axis}[
              xbar,
              font = \itshape,
              y axis line style = { opacity = 0 },
              axis x line       = none,
              tickwidth         = 0pt,
              width 			  = 11cm,
              ytick 			  = data,
              enlarge y limits  = 0.2,
              enlarge x limits  = 0.12,
              symbolic y coords = {Reference Work; Empirical Results,Tutorial; Empirical Results,Reference Work; Software Manual,Survey; Resource,Tutorial; Reference Work; Software Manual,Survey; Tutorial,Survey; Reference Work,Tutorial; Reference Work,Tutorial; Software Manual,Survey; Empirical Results,Tutorial; Resource; Software Manual,Resource; Software Manual,Resource; Empirical Results},
              nodes near coords
              ]
			\addplot coordinates {(30,Resource; Empirical Results) (14,Resource; Software Manual) (11,Tutorial; Resource; Software Manual) (10,Survey; Empirical Results) (8,Tutorial; Software Manual) (4,Tutorial; Reference Work) (3,Survey; Reference Work) (2,Survey; Tutorial) (1,Tutorial; Reference Work; Software Manual) (1,Survey; Resource) (1,Reference Work; Software Manual) (1,Tutorial; Empirical Results) (1,Reference Work; Empirical Results)};
			\end{axis}
		\end{tikzpicture}
		\caption{Distribution of pedagogical roles for documents in full annotated corpus with more than one role}
		\label{fig:multipleroles}
\end{figure*}

\begin{table*}[bht]
\centering
\def\arraystretch{1.2}
\begin{tabular}{c S[table-format=3] p{10cm}}
\hline
Pedagogical role & {Cluster ID} & Example sentence \\
\hline\hline
\textit{Survey} & 250 & This view has been worked out in the text generation and dialog community more than in the text understanding community (Mann and Thompson, 1987; Hovy, 1993; Moore, 1994). \\
& 123 & Confronted with the claim that Game Theory should be the theoretical backbone to NLG, some people might respond that no new backbone is needed, because the theory of formal languages, conjoined with a properly expressive variant of Symbolic Logic, provides sufficient backbone already. \\ \hline

\textit{Tutorial} & 209 & As you guessed from my explanations of different notations, different regex engine designers unfortunately have different ideas about the syntax to use. \\
& 95 & This information is incorporated in the tri-factorization model via a squared loss term, where the notation Tr (4) means trace of the matrix A. \\ \hline

\textit{Resource} & 147 & {\tt >\kern.1pt>\kern.1pt> windowdiff(s1, s1, 3)} \\
& 255 & ...     {\tt print(\textquotesingle\textquotesingle, repr(corpus.fileids())[:60])} \\ \hline

\textit{Reference Work} & 155 & The greater the resumption of the activity (i.e., mismatch negativity), the more different the neurological processing of the new item. \\
& 86 & A trajectory of an object is determined by its different centers of gravity relative to an underlying coordinate system. \\ \hline

\textit{Empirical Results} & 183 & 5.3 Using Multiple Knowledge Sources \\
& 62 & The NCC open track is shown in the following table 2.\\ \hline

\textit{Software Manual} & 147 & {\tt >\kern.1pt>\kern.1pt> clf.fit(X, Y)} \\
& 152 & An example of this approach can be found in the /verbi folder in the Italian MOR grammar. \\ \hline

\textit{Other} & 279 & The problem in the cases (3) and (4) is how and why the hearer fails to derive implicatures. \\
& 157 & Proofs of the form suppose-absurd F D are called proofs by contradiction. \\
\hline\hline
\end{tabular}

\caption{Example sentences from the clusters most frequently associated with each pedagogical role. The clusters representing mostly punctuation, numbers, or incoherent strings were not included in calculating most frequently associated clusters.}
\label{table:roleexamples}
\end{table*}

\end{document}